\documentclass[10pt,twocolumn,letterpaper]{article}

\usepackage[pagenumbers]{wacv} 

\usepackage{graphicx}
\usepackage{amsmath}
\usepackage{amssymb}
\usepackage{booktabs}

\usepackage[pagebackref,breaklinks,colorlinks]{hyperref}

\usepackage[capitalize]{cleveref}
\crefname{section}{Sec.}{Secs.}
\Crefname{section}{Section}{Sections}
\Crefname{table}{Table}{Tables}
\crefname{table}{Tab.}{Tabs.}

\def\wacvPaperID{470} 
\def\confName{WACV}
\def\confYear{2025}

\begin{document}

\def\modelname{\textbf{VipDiff}\xspace}
\title{\modelname: Towards Coherent and Diverse Video Inpainting via Training-free Denoising Diffusion Models}

\author{
Chaohao Xie \qquad Kai Han\thanks{~Corresponding author} \qquad Kwan-Yee K. Wong\footnotemark[1] \\
The University of Hong Kong \\
\tt\small{viousxie@connect.hku.hk}, \tt\small{kaihanx@hku.hk}, \tt\small{kykwong@cs.hku.hk}
}


\maketitle

\begin{abstract}
Recent video inpainting methods have achieved encouraging improvements by leveraging optical flow to guide pixel propagation from reference frames, either in the image space or feature space. However, they would produce severe artifacts when the masked area is too large and no pixel correspondences could be found. Recently, denoising diffusion models have demonstrated impressive performance in generating diverse and high-quality images, and have been exploited in a number of works for image inpainting. These methods, however, cannot be applied directly to videos to produce temporal-coherent inpainting results. In this paper, we propose a training-free framework, named \modelname, for conditioning diffusion model on the reverse diffusion process to produce temporal-coherent inpainting results without requiring any training data or fine-tuning the pre-trained models. \modelname takes optical flow as guidance to extract valid pixels from reference frames to serve as constraints in optimizing the randomly sampled Gaussian noise, and uses the generated results for further pixel propagation and conditional generation. \modelname also allows for generating diverse video inpainting results over different sampled noise. Experiments demonstrate that our \modelname outperforms state-of-the-art methods in terms of both spatial-temporal coherence and fidelity.
\end{abstract}

\section{Introduction}
\label{sec:intro}

Video inpainting aims to generate spatial-temporal coherent contents for the masked areas in corrupted video frames. Existing video inpainting methods can be broadly classified into 1) end-to-end synthesis methods and 2) flow-based pixel propagation methods.

\begin{figure*}[hbt]
	\small
	\setlength{\tabcolsep}{1.0pt}
	\centering
	\begin{tabular}{ccccccc}
		\includegraphics[width=24.4mm]{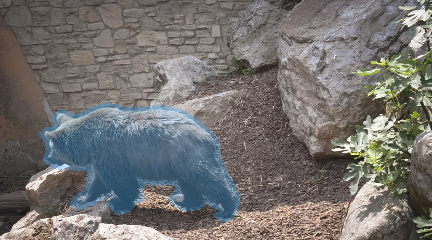} &				
		\includegraphics[width=24.4mm]{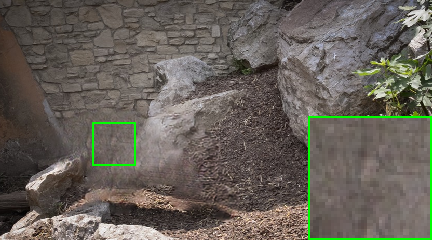} &				
		\includegraphics[width=24.4mm]{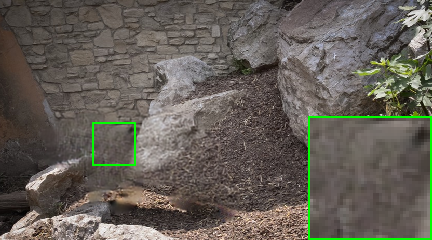} &				
		\includegraphics[width=24.4mm]{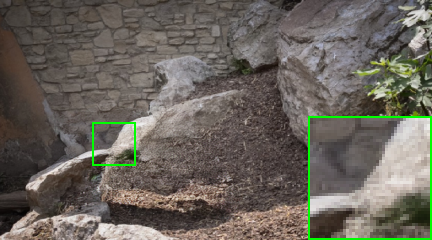} &				
		\includegraphics[width=24.4mm]{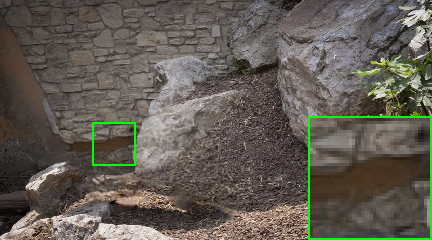} &				
		\includegraphics[width=24.4mm]{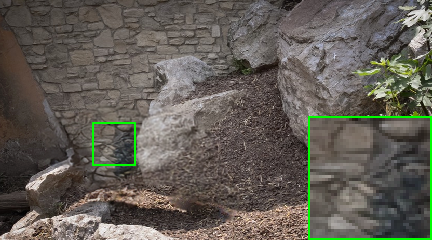} &				
		\includegraphics[width=24.4mm]{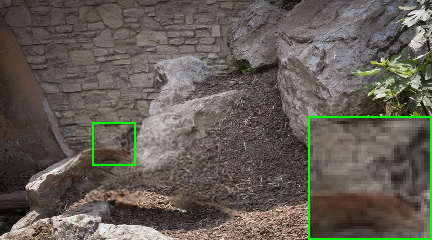} \\

		\includegraphics[width=24.4mm]{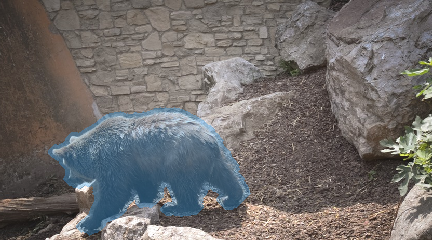} &				
		\includegraphics[width=24.4mm]{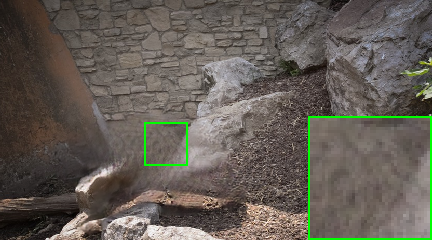} &				
		\includegraphics[width=24.4mm]{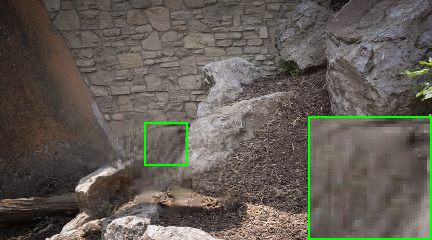} &				
		\includegraphics[width=24.4mm]{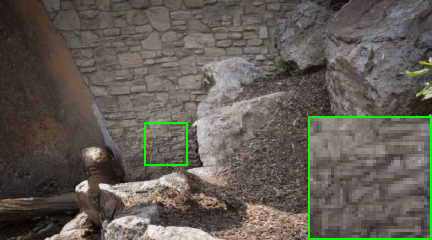} &				
		\includegraphics[width=24.4mm]{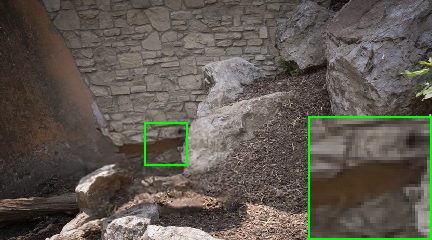} &				
		\includegraphics[width=24.4mm]{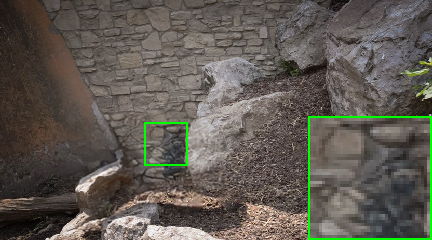} &				
		\includegraphics[width=24.4mm]{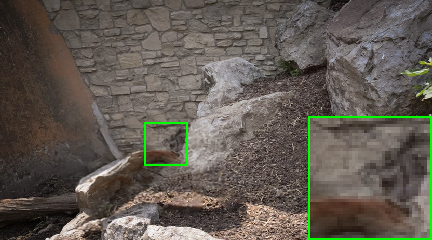} \\

		\includegraphics[width=24.4mm]{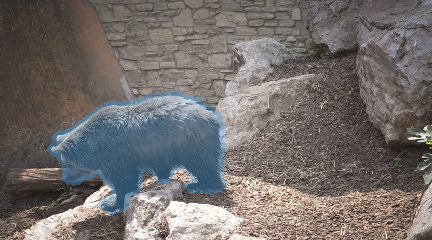} &				
		\includegraphics[width=24.4mm]{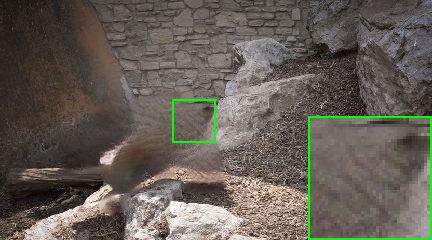} &				
		\includegraphics[width=24.4mm]{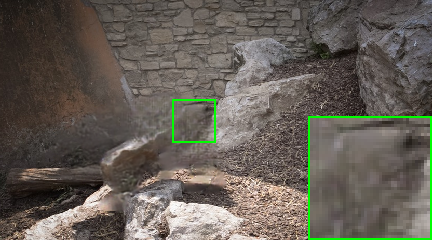} &				
		\includegraphics[width=24.4mm]{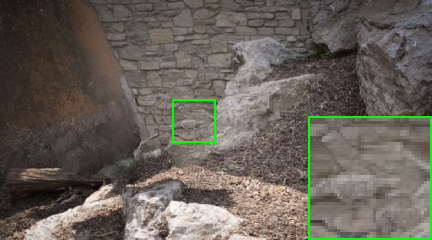} &				
		\includegraphics[width=24.4mm]{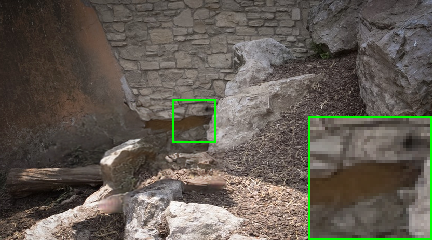} &				
		\includegraphics[width=24.4mm]{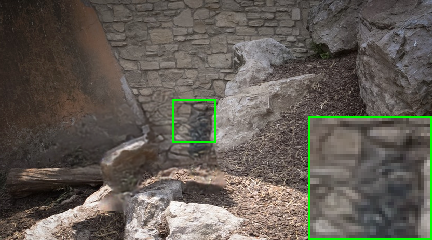} &				
		\includegraphics[width=24.4mm]{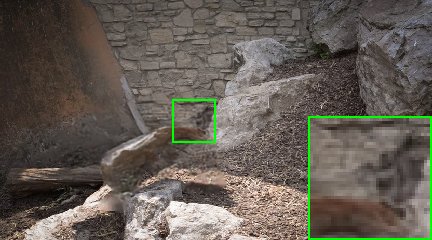} \\

		\includegraphics[width=24.4mm]{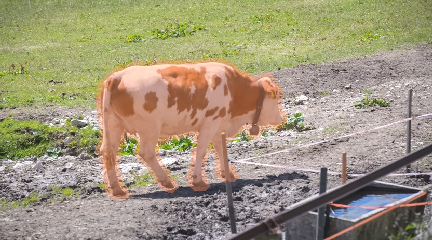} &				
		\includegraphics[width=24.4mm]{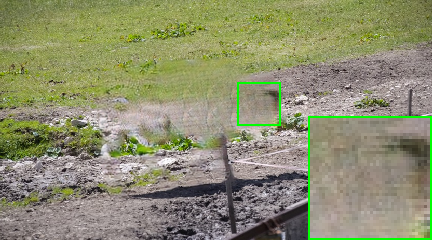} &				
		\includegraphics[width=24.4mm]{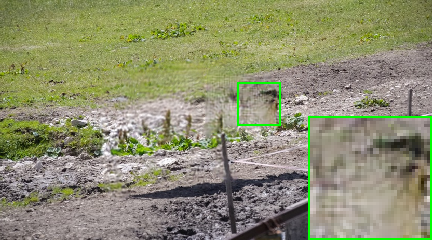} &				
		\includegraphics[width=24.4mm]{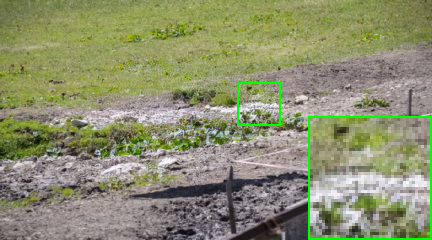} &				
		\includegraphics[width=24.4mm]{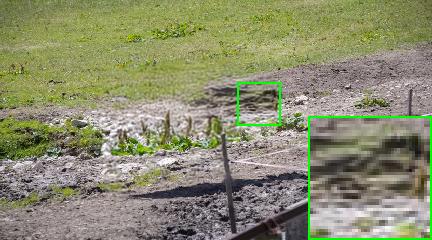} &				
		\includegraphics[width=24.4mm]{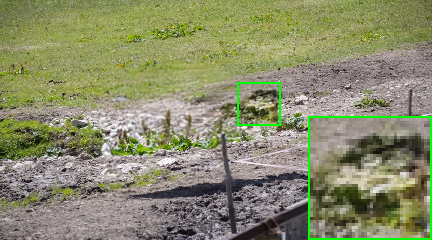} &				
		\includegraphics[width=24.4mm]{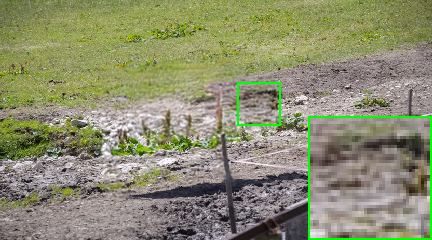} \\				

		\includegraphics[width=24.4mm]{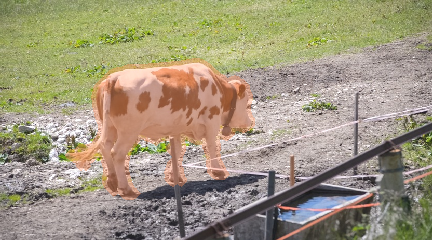} &				
		\includegraphics[width=24.4mm]{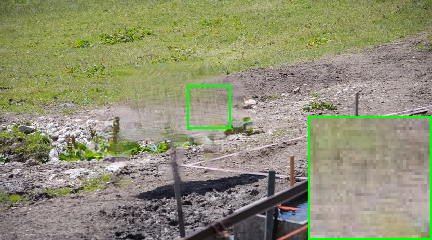} &				
		\includegraphics[width=24.4mm]{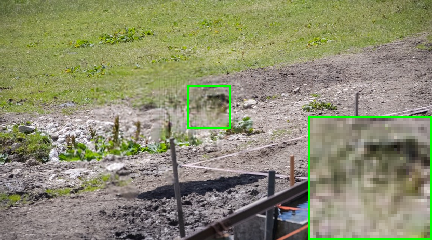} &				
		\includegraphics[width=24.4mm]{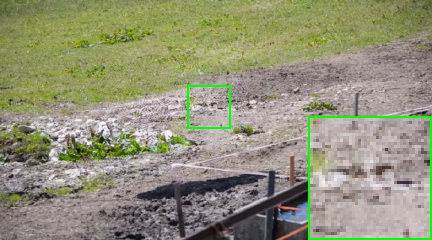} &				
		\includegraphics[width=24.4mm]{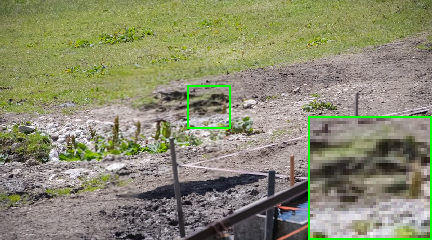} &				
		\includegraphics[width=24.4mm]{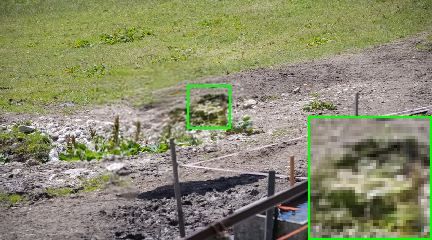} &				
		\includegraphics[width=24.4mm]{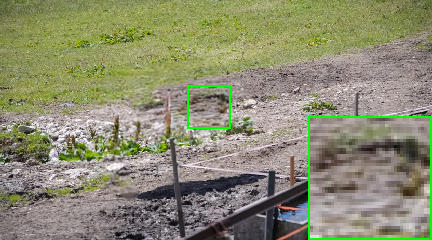} \\		

		\includegraphics[width=24.4mm]{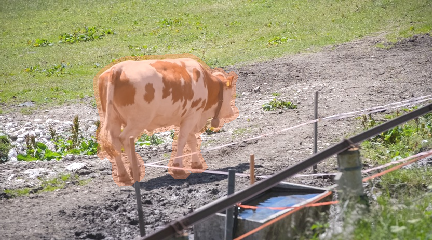} &				
		\includegraphics[width=24.4mm]{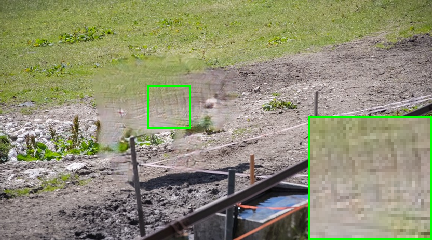} &				
		\includegraphics[width=24.4mm]{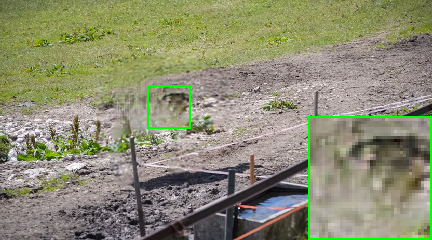} &				
		\includegraphics[width=24.4mm]{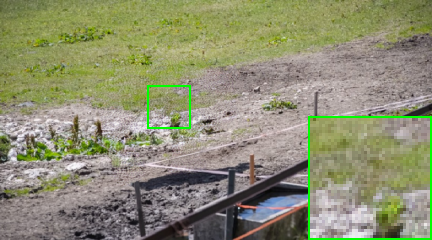} &				
		\includegraphics[width=24.4mm]{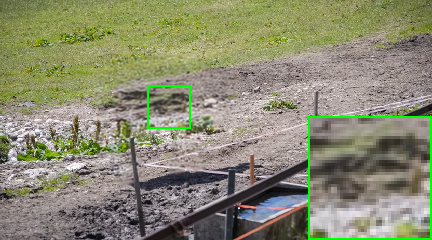} &				
		\includegraphics[width=24.4mm]{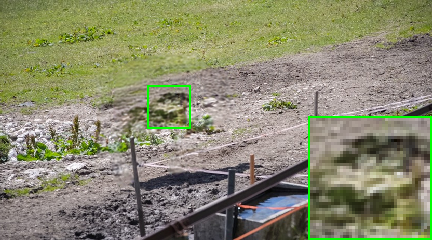} &				
		\includegraphics[width=24.4mm]{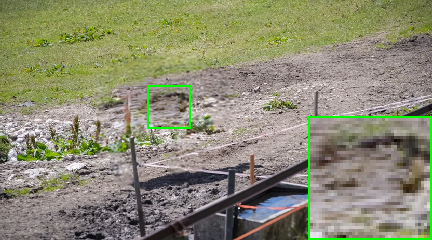} \\

		(a) Input & (b) ProPainter~\cite{zhou2023propainter} & (c) ECFVI~\cite{kang2022error} & (d) LDM~\cite{rombach2022high} & (e) Ours-Sample1 & (f) Ours-Sample2 &  (g) Ours-Sample3 \\

	\end{tabular}
        \vspace{-1.0em}
	\caption{Video inpainting results on Davis dataset, we randomly select 3 frames from 'bear' and 'cows' videos. LDM generates different contents, ProPainter and ECFVI contain artifacts in the mask center, while our samples are more sharp and temporal-coherent. 
 }
	\label{fig1_compare}

\vspace{-1.5em}
\end{figure*}

End-to-end synthesis methods typically adopt 3D convolutions~\cite{chang2019learnable,chang2019free,hu2020proposal}, attention modules~\cite{lee2019copy,li2020short,zeng2020learning}, or multi-frame transformer-based networks~\cite{zeng2020learning,liu2021fuseformer,li2022towards,zhou2023propainter}. They take corrupted frames and their corresponding masks as input, and train a video completion network to directly output the completed frames. Since they have fixed temporal receptive windows (i.e., only a fixed length of reference frames can be fed to the input), severe artifacts and temporal incoherence often occur when the input frames fail to provide useful texture and content hints for the frame completion models (see Fig.~\ref{fig1_compare}(b)). 

Flow-based pixel propagation methods first conduct optical flow completion and then utilize the completed flows to guide a pixel propagation step, which warps valid pixels from reference frames to a target frame. An additional synthesis network is trained to fill the remaining masked pixels. Although these flow-based methods~\cite{xu2019deep,gao2020flow,zhang2019internal,kang2022error} can achieve better temporal coherence over the generated frames, their results still show obvious artifacts in the mask center when the masks are large (see Fig.~\ref{fig1_compare}(c)). Undoubtedly, there is still a large room for improvement.

Recently, denoising diffusion models~\cite{sohl2015deep,ho2020denoising} have attracted enormous attention due to their impressive performance in generating diverse and high-quality images from Gaussian noise through a series of denoising steps. There are several methods, like DPS~\cite{chung2023diffusion} and ReSampling~\cite{trippe2023diffusion}, which adopt pre-trained diffusion model for image inpainting via sampling posterior distributions. RePaint~\cite{lugmayr2022repaint} introduces a time travel strategy to improve spatial coherence between masked and unmasked areas, and CoPaint~\cite{zhang2023coherent} proposes optimizing the randomly sampled Gaussian noise using the unmasked pixels as constraints. Latent Diffusion Models (LDM)~\cite{rombach2022high} map the input to a low-dimensional space with an encoder and feed the encoded features for conditional generation in the latent feature space. These diffusion-based methods work well for the image inpainting task. However, they cannot be applied directly to videos since they do not consider temporal coherence. For instance, we show the inpainting results of LDM on three frames randomly selected from two videos in Fig.~\ref{fig1_compare}(d). Even though the surrounding patterns are similar, LDM generates different contents in the masked area in different frames and fails to maintain temporal coherence.

Inspired by the above diffusion-based image inpainting methods, we propose a training-free framework for conditioning pre-trained image-level diffusion models to generate spatial-temporal coherent and diverse video inpainting results. Given a corrupted video, we first adopt a flow completion model to predict all the optical flows between frames. We then use the completed flows to propagate valid pixels from reference frames to a target frame. Next, we carry out the reverse diffusion process, where valid pixels (i.e., both pixels in the unmasked area and pixels propagated from other frames) in the target frame serve as constraints in optimizing the sampled noise through backpropagation. 

After optimization, a spatial-temporal coherent frame is generated. Its pixels can be propagated to other frames again and the process is repeated for the next target frame.
Note that once we generate one completed frame and propagate its pixels to other frames, the number of remaining invalid (masked) pixels in most frames, especially those nearby ones, would become zero. Since we only need to perform conditional generation on those frames with a non-zero size mask, our framework is therefore very time efficient. \modelname can generate both spatial-temporal coherent results and diverse inpainted videos. In Fig.~\ref{fig1_compare}(e), (f) and (g), we show three different results on two videos. All of our generated results are coherent with the video motions.

\modelname has the following advantages. 1) It eliminates the efforts in collecting large-scale video datasets and training a large video diffusion model, which may only be possible to be conducted by large companies or foundations. 2) It leverages the diverse generation capability of pre-trained image-level diffusion models and generates results far better than SOTA methods, while still allows users to choose from diverse video inpainting results. 3) It just takes several minutes to inpaint a video sequence of about 100 to 200 frames on a single RTX 3090 GPU, demonstrating that it is both device friendly and time efficient. Hence our method enables users with limited GPU resources to utilize the ability of pre-trained diffusion models for video inpainting.

We summarize our contributions as follows:
\begin{itemize}
    \item A training-free framework, named \modelname, for conditioning pre-trained image-level diffusion model to generate spatial-temporal coherent video inpainting results. On large mask cases, it is capable of generating diverse results for users.
    \item \modelname saves the extensive computation and time costs in specifically training a video denoising diffusion model for video inpainting task.
    \item To the best of our knowledge, we present the first work that successfully tames pre-trained image-level diffusion models for video inpainting task.
    \item Experiments on public datasets demonstrate superior performance of \modelname over existing state-of-the-arts.
\end{itemize}

\section{Related Works}

In this section, we give a brief review of video inpainting methods, including recent methods based on diffusion probabilistic model~\cite{sohl2015deep}. 

\subsection{Patch-based Methods} Traditional inpainting methods~\cite{Barnes:2009:PAR} analyze local structures and fluid dynamics~\cite{bertalmio2001navier} along the mask boundary, and exploit spatial-temporal similarities~\cite{patwardhan2005video,wexler2007space} between patches from undamaged areas to sample patches for filling the masked regions. To improve the performance on scenes with occluded objects, additional energy functions~\cite{patwardhan2007video,granados2012not,newson2014video} and optical-flow constraints ~\cite{matsushita2006full,huang2016temporally} have been proposed to guide the optimization process. Although these methods can produce plausible results, their lack of understanding of global semantics results in poor visual quality for complex scenes. They also require substantial amount of time for computation, which limits their potentials in practical applications.

\subsection{Deep Learning Methods} With the great progress in deep learning, deep neural networks for video inpainting have become popular, and large performance gains have been reported. To capture spatial-temporal relationships among masked frames, 3D convolutions have been widely adopted to combine temporal information and local spatial structures~\cite{chang2019learnable,wang2019video,chang2019free,hu2020proposal}. To improve visual coherence over the generated sequences, optical flows have been utilized to guide the propagation of pixels from nearby frames into the masked areas as additional prior~\cite{xu2019deep,gao2020flow,zhang2019internal}. To obtain optical flows in the masked areas, an additional flow completion network is trained to complete the global flows. ECFVI~\cite{kang2022error} introduces an error correcting model for reducing color discrepancy when propagating pixels from far neighbors. Other methods~\cite{lee2019copy,li2020short,zeng2020learning} utilize different attention modules for aggregating valid features from different frames.  

Recently, Transformer~\cite{vaswani2017attention} and Vision Transformer~\cite{vit_dosvit} have shown great potential in video inpainting. STTN~\cite{zeng2020learning} introduces spatial-temporal transformer to combine multi-scale information across different frames and fill masked areas in all input frames simultaneously. FuseFormer~\cite{liu2021fuseformer} adopts overlapping patch splitting strategy for learning fine-grained features to enhance inpainting quality. FAST~\cite{Yu_2021_ICCV} combines information from the frequency domain. E2FGVI~\cite{li2022towards}, FGT~\cite{zhang2022flow,zhang2023exploiting} and ProPainter~\cite{zhou2023propainter} incorporate optical flows in training the transformer structures. Although these transformer-based methods improve the overall visual quality, they still suffer from performance degradation when facing larger masks or when neighbouring frames fail to provide useful information.

\subsection{Diffusion-based Methods} More recently, diffusion-based methods~\cite{sohl2015deep,ho2020denoising,song2021denoising} have made significant progress in image generation. To generate coherent inpainted images, DPS~\cite{chung2023diffusion} and ReSampling~\cite{trippe2023diffusion} treat the image inpainting problem as a process of sampling images from the posterior distributions conditioned on the masked images. RePaint~\cite{lugmayr2022repaint} adopts a time travel strategy which combines the current step $t$ denoising result with the noise degraded image based on the mask and uses it to generate the step $t + 1$ image by a one-step forward process. CoPaint~\cite{zhang2023coherent} improves the spatial coherence by taking the unmasked pixels as constraints and optimizing the noise $z$ to match the generated image with the masked image. Latent Diffusion Models (LDM)~\cite{rombach2022high} utilize an auto-encoder structure to project the masked image into a latent space with reduced resolution for conditional generation. These methods can produce diverse and high-quality inpainted images. However, they cannot be directly applied to video inpainting, since they only consider spatial coherence within an image but not temporal coherence among frames. Other methods try to train large video diffusion models for video sequences editing~\cite{zhang2023avid,gu2023flowguided}, or combining language models with text prompts~\cite{wu2024lgvi} to act as agents. Even trained with large models, they do not explicitly address the temporal inconsistent issues. Our proposed method combines the advantages of diffusion-based methods and flow-based methods, and generates spatial-temporal coherent and diverse results for video inpainting, which only requires on image-level pre-trained diffusion models.

\begin{figure*}[hbt]
    \centering
    \includegraphics[width=0.99\textwidth]{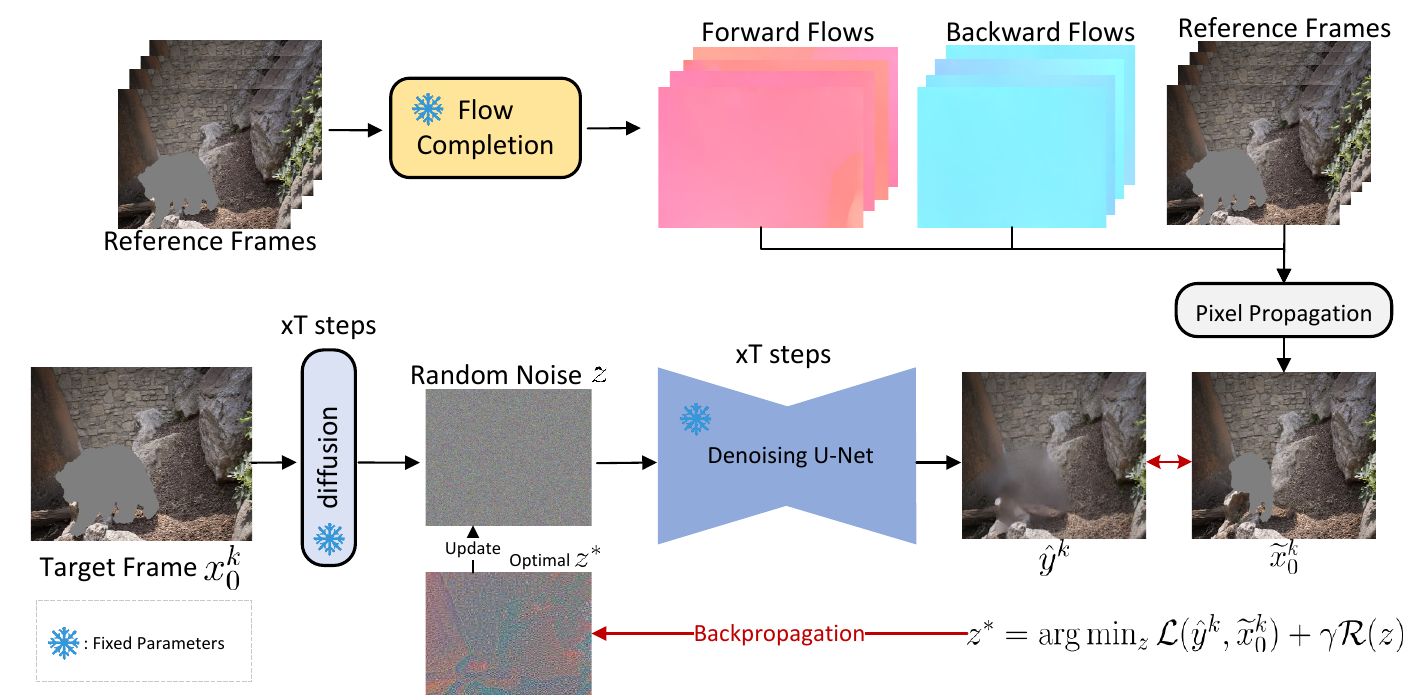}
    \caption{Overall framework of our \modelname. Given a target frame $x^{k}_{0}$, we first adopt a flow completion model to predict the optical flows. Then the flows are utilized for pixel propagation to extract temporal prior from reference frames to get partially inpainted image $\widetilde{x}^{k}_{0}$. Next, $\widetilde{x}^{k}_{0}$ will act as constrains for optimizing the random sampled Gaussian nosie $z$ which is feed for the reverse denoising U-Net. Through backpropagation, we optimize the noise $z$ at each time step and finally find an optimal $z^{*}$ for filling the target frame. Note that all the model parameters are frozen during the noise optimization process. }
    \label{generation_framework}
\vspace{-1.5em}
\end{figure*}

\section{Proposed Methods}
In this section, we first give a brief overview of the Denoising Diffusion Probabilistic Models (DDPMs) and describe the notations used in this paper. We then introduce our training-free framework for conditioning pre-trained image-level diffusion models to generate spatial-temporal coherent video inpainting results. We also recommend readers to refer to the original papers~\cite{sohl2015deep,ho2020denoising} of the diffusion models for details and derivations if needed. 

Let $ \mathcal{X} = \{ {x}^{k} | k \in [1, N] \}$ denote a corrupted video with $N$ frames, and $\mathcal{M} = \{ m^{k} | k \in [1, N] \} $ be the corresponding per-frame masks with the same spatial dimensions as $ \mathcal{X} $. A video inpainting method takes the image frames $\mathcal{X}$ and the masks $\mathcal{M}$ as input, examines the frame core, and outputs the recovered video $\hat{\mathcal{Y}} = \{ \hat{y}^{k} | k \in [1,N] \}$, which should be spatial-temporal coherent.

\subsection{Preliminaries}
Denoising diffusion probabilistic models (DDPMs)~\cite{ho2020denoising} can be viewed as a type of generative models which are trained to approximate a data distribution $x_0 \sim p(x)$ through a series of intermediate variables $x_{1:T}$ that are degraded states of $x_0$. These models comprise two processes, namely the {\em forward diffusion process} which defines how the original image $x_0$ becomes the noise image $x_T$, and the {\em reverse diffusion process} which recovers the clean image $x_0$.

Given the $k$-th masked input frame $x^{k}_{0} = x^{k} * (1 - m^{k})$ from the corrupted video $\mathcal{X}$, the forward diffusion process defines a Markov chain $q$ of $T$ steps which progressively adds Gaussian noise to the damaged image $x^{k}_{0}$ by a variance schedule $ \{ \beta_{t} \}^{T}_{t=1}$, i.e.,
\begin{equation}\label{forward_diffusion}
    q(x^{k}_{t} | x^{k}_{t - 1}) = \mathcal{N} (x^{k}_{t}; \sqrt{1 - \beta_{t}} x^{k}_{t - 1}, \beta_{t} \mathbf{I}) 
\end{equation}
where each $\beta_{t} \in (0,1)$. With a sufficiently large $T$, the final state $q(x^{k}_{T} | x^{k}_{0})$ will be close to real Gaussian noise. 

For the reverse diffusion process, a diffusion model with learnable parameters $\theta$ is adopted to estimate the amount of noise at each time step ($x^{k}_{t}$),
\begin{equation}\label{reverse_diffusion}
    p_{\theta}(x^{k}_{t - 1} | x^{k}_{t}) = \mathcal{N} (x^{k}_{t - 1}; \mu_{\theta}(x^{k}_{t}, t), \Sigma_{\theta}(x^{k}_{t}, t)).
\end{equation}

To perform unconditional generation from a sampled Gaussian noise $x^{k}_{T}$, a shared-weight denoising auto-encoder $\epsilon_{\theta} (x^{k}_{t}, t)$ is trained to predict the denoised version of $x^{k}_{t}$. The objective can be formulated as
\begin{equation}\label{ddpm_objective}
    \mathcal{L}_{DDPM} = \mathbb{E}_ { x^{k}_{0}, \epsilon \sim \mathcal{N}(\mathbf{0}, \mathbf{I}), t \sim \mathcal{U} (1, T) } [ \| \epsilon - \epsilon_{\theta} (x^{k}_{t}, t) \|^{2} ],
\end{equation}
where $\mathcal{U} (1, T)$ is a uniform distribution on $\{ 1,2,..., T \}$ and $\epsilon$ is the noise level at current time step.

Latent Diffusion Models (LDMs)~\cite{rombach2022high} are variants of DDPMs which utilize an encoder $\mathcal{E}$ to project the input image $x^{k}$ into low-resolution latent representations 
and a decoder $\mathcal{D}$ to reconstruct the image. The forward diffusion process and reverse diffusion process are carried out in the latent space. LDMs allow for conditional generation by introducing additional prior in the reverse process $p_{\theta}(x^{k}_{t - 1} | x^{k}_{t}, c)$, where $c$ denotes the condition which can be texts, object segmentation mask, or other useful information. For inpainting, the condition $c$ would be provided by the masked input frame $x_0^{k}$. The denoising network $\epsilon_{\theta} (x^{k}_{t}, t, c)$ can be trained under conditional reverse process. 

Note that conditional LDMs only inpaint a single frame at a time without enforcing any temporal coherence between frames. Na\"ively applying conditional LDMs to videos will inevitably lead to temporal incoherence in the generated sequence (see Fig.~\ref{fig1_compare}). In this paper, we propose a training-free framework that successfully tames image-level diffusion models for generating spatial-temporal coherent inpainting results, by utilizing optical flow to induce prior from pixels in nearby frames.

\begin{table*}[t]
	\centering
	\caption{Quantitative comparison with state-of-the-art video inpainting methods on the YouTube-VOS and DAVIS datasets. $\uparrow$ ($\downarrow$) indicates the higher (lower) the better. We highlight the best results in \textbf{bold}, and the second best in \underline{underline} . }
	\label{metric_compare}
	\begin{tabular}{c|cccc||cccc||c}
		\hline  
		
		\hline

         & \multicolumn{4}{c||}{YouTube-VOS} & \multicolumn{4}{c||}{DAVIS} & Speed \\
        \cline{2-10}

        Methods & PSNR$\uparrow$ & SSIM$\uparrow$ & VFID$\downarrow$ & $\text{E}_{warp}$ $\downarrow$ & PSNR$\uparrow$ & SSIM$\uparrow$ & VFID$\downarrow$ & $\text{E}_{warp}$ $\downarrow$ & s/frame\\ 
  
	\hline

        \hline
        LGTSM~\cite{chang2019learnable} & 29.74 & 0.9504 & 0.070 & 0.1859 & 28.57 & 0.9409 & 0.170 & 0.1640 & 0.19\\

        VINet~\cite{kim2019deep} & 29.20 & 0.9434 & 0.072 & 0.1490 & 28.96 & 0.9411 & 0.199 & 0.1785 & - \\

        DFVI~\cite{xu2019deep} & 29.16 & 0.9429 & 0.066 & 0.1509 & 28.81 & 0.9404 & 0.187 & 0.1608 & 1.96\\

        CAP~\cite{lee2019copy} & 31.58 & 0.9607 & 0.071 & 0.1470 & 30.28 & 0.9521 & 0.182 & 0.1533 &  0.27 \\

        FGVC~\cite{gao2020flow} & 29.67 & 0.9403 & 0.064 & 0.1022 & 30.80 & 0.9497 & 0.165 & 0.1586 & 1.75 \\

        STTN~\cite{zeng2020learning} & 32.34 & 0.9655 & 0.053 & 0.0907 & 30.67 & 0.9560 & 0.149 & 0.1449 & 0.06\\

        FuseFormer~\cite{liu2021fuseformer} & 33.29 & 0.9681 & 0.053 & 0.0900 & 32.54 & 0.9700 & 0.138 & 0.1362 & 0.15\\

        E2FGVI~\cite{li2022towards} & 33.71 & 0.9700 &  \underline{0.046} & \underline{0.0864} & 33.01 & 0.9721 & 0.116 & 0.1315 & 0.13 \\

        FGT~\cite{zhang2022flow} & 33.81 & {0.9711} &  0.066 &   0.0893 & 33.79 & \underline{0.9737}  & 0.106 & 0.1331 
 & 0.79\\

        ECFVI~\cite{kang2022error} & 33.77 &  0.9703 & 0.053 & 0.0913  & 33.52 & 0.9732 & 0.105 & \underline{0.1292} & 1.34 \\

        ProPainter~\cite{zhou2023propainter} & \textbf{34.23} & \underline{0.9764} & 0.054 & 0.0971 & \textbf{34.27} & 0.9731 & \underline{0.104} & 0.1312 & 0.09 \\

        \hline

        Ours & \underline{34.21} & \textbf{0.9773}  & \textbf{0.041} & \textbf{0.0828} & \underline{34.23} & \textbf{0.9745} & \textbf{0.102} & \textbf{0.1280} & 2.73\\ 
		
		\hline
		
		\hline
		
	\end{tabular} 

\vspace{-1.5em}
\end{table*}

\subsection{Optical Flow Guided Pixel Propagation}
Since pre-trained models~\cite{diffusionbeatsgan} or LDMs~\cite{rombach2022high} have demonstrated strong capability in producing high-quality inpainting results from masked input images, we propose to directly make use of the them in our method without modifying any of its network structure (attention modules, etc.) or fine-tuning its parameters. Unlike other methods which train a video diffusion model for specific video generation tasks, we optimize the randomly sampled Gaussian noise $z$ to maximize the spatial-temporal coherence by leveraging both the unmasked pixels in the target frame and pixels propagated from reference frames as constraints. 

Figure.\ref{generation_framework} shows the overall generation pipeline of our \modelname, the key to making the inpainted frames temporal-coherent lies in the success of deriving proper prior for the masked areas from as many valid pixels from nearby frames as possible. To this end, we adopt a pre-trained RAFT model~\cite{teed2020raft} as our flow estimator. Pre-trained flow completion RAFT is capable of predicting complete optical flows for images with missing areas. The completed flow from frame $k$ to frame $j$ is denoted as
\begin{equation}\label{flow_complete}
    \widetilde{f}_{k\to j} = F( x^{k}_{0}, x^{j}_{0}, m^{k}, m^{j} ),
\end{equation}
where $F(\cdot)$ is the flow estimator. After obtaining $\widetilde{f}_{k\to j}$, we can propagate valid pixels from frame $j$ (referred to as the {\em reference frame}) to frame $k$ (referred to as the {\em target frame}) to provide temporal constraints for the diffusion model.

\subsubsection{Pixel Propagation}
Let $x^{j}_{0}$ be a source frame of $x^{k}_{0}$ that contains valid pixel information for the masked area of $x^{k}_{0}$. We define a backward warping function $\omega(\cdot)$ to propagate valid pixels from the source frame $x^{j}_{0}$ to the masked area of the target frame $x^{k}_{0}$ based on the completed flow $\widetilde{f}_{k\to j}$. Written as
\begin{equation}\label{propagate_pixels}
\begin{aligned}
    & \widetilde{x}^{k}_{0} = x^{k}_{0} + m^{j \to k } \odot \omega(x^{j}_{0}, \widetilde{f}_{k\to j}), \\
    & m^{j \to k } = m^{k} \odot (1 - \omega (m^{j}, \widetilde{f}_{k\to j})  ),
\end{aligned}
\end{equation}
where $ m^{j \to k }$ denotes a mask for the propagated valid pixels falling within the masked area of $x_0^{k}$. After pixel propagation, we update the invalid mask of $x_0^{k}$ to $\widetilde{m}^{k} = m^{k} - m^{j \to k}$. We repeat the above pixel propagation step for different $j$, starting from the nearest to the furthest neighboring frame, until the invalid mask becomes all zeros or all neighboring frames have been exhausted.
As pointed out in~\cite{kang2022error}, directly propagating pixels from other frames may result in color discrepancy due to misalignment or brightness inconsistency. The authors introduced an error compensation network to reduce this problem. We adopt the same error compensation network in our work to reduce the color discrepancy in pixel propagation. 

\begin{figure}[!bt]
    \centering
    \includegraphics[width=0.46\textwidth]{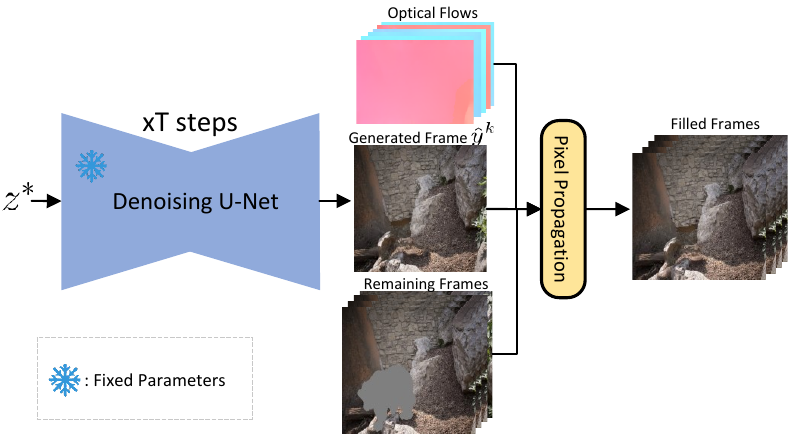}
    \caption{Process for propagating constrained generated pixels to other frames.}
    \label{to_entire_sequence}

\vspace{-1.5em}
\end{figure}

\subsection{Noise Optimized Reverse Diffusion Process}
After applying the above optical flow guided pixel propagation, we obtain a (partially) filled frame $\widetilde{x}^{k}_{0}$ which contains fewer invalid pixels compared with ${x}^{k}_{0}$. We can then proceed to generate an inpainted frame $\hat{y}^k$ for $\widetilde{x}^k_0$ based on the reverse diffusion process. The goal is to sample a Gaussian noise $z$ that maximizes the similarity between the generated image $\hat{y}^k$ and $\widetilde{x}^k_0$. This can be achieved by optimizing the noize $z$ through backpropagation with the following objective

\begin{equation}\label{cond_loss}
    \mathcal{L}_{cond}  = \| \hat{y}^{k} \odot (1 - \widetilde{m}^{k})  - \widetilde{x}^{k}_{0} \odot (1 - \widetilde{m}^{k}) \|^{2}.
\end{equation}
Specifically, $\hat{y}^{k} = \epsilon^d_\theta(z, t, {x}^{k}_{0})$ with $\epsilon^d_\theta$ being the pre-trained latent diffusion model. In general, our method can work for unconditional diffusion models with  $\hat{y}^{k} = \epsilon^d_\theta(t, {x}^{k}_{t})$. Following~\cite{zhang2023coherent}, we also add a regularization term $\mathcal{R}(z)=\| z - z_0\|^{2}$ to constrain $z$ with the original sampled noise $z_0$ to follow a Gaussian distribution. Hence, the overall optimization objective can be written as
\begin{equation}
    z^{*} = \arg\min_{z} \mathcal{L}_{cond} (\hat{y}^{k}, \widetilde{x}^{k}_{0}) + \gamma\mathcal{R}(z),
\end{equation}
where $\gamma$ is the weight of the regularization term.

With the optimal $z^{*}$, we can generate a spatial-temporal coherent inpainting result for the target frame. The spatial coherence can be achieved by the valid pixels existing in the target frame, and the temporal consistency comes from the constraints of propagated pixels. The pre-trained diffusion models provide overall coherence for the unconstrained pixels. Note that we do not tune any parameters of the pre-trained diffusion model but only optimize $z$ throughout the entire process, making our method device friendly and time efficient. The generation framework is shown in Fig.~\ref{generation_framework}. 

\vspace{-1.0em}
\subsubsection{Inpaint the Entire Sequence} After obtaining the inpainted frame $\hat{y}^{k}$, we update its invalid mask to all zeros. We then proceed to the next frame and, likewise, carry out optical flow guided pixel propagation. Note that once we have a completed frame with no invalid pixels, in most cases, many subsequent frames can already be completed by pixel propagation alone (see Fig.~\ref{to_entire_sequence}). We only need to carry out the noise optimized reverse diffusion process for those frames which remain incomplete after pixel propagation (the number of such frames is usually relatively small compared to the video length). The whole process is repeated for each frame in the sequence until all frames have been completed.

\begin{figure*}[hbt]
	\small
	\setlength{\tabcolsep}{1.0pt}
	\centering
	\begin{tabular}{ccccc}
 
		\includegraphics[width=34.15mm]{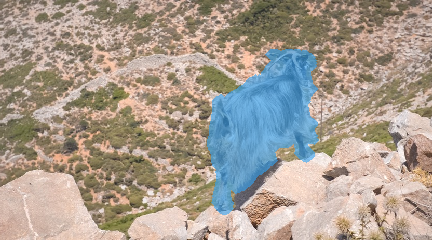} &
		\includegraphics[width=34.15mm]{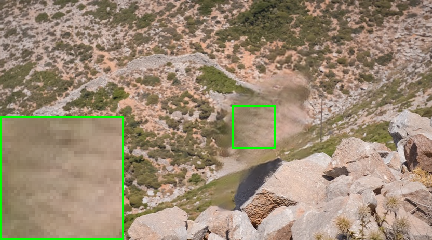} &
		\includegraphics[width=34.15mm]{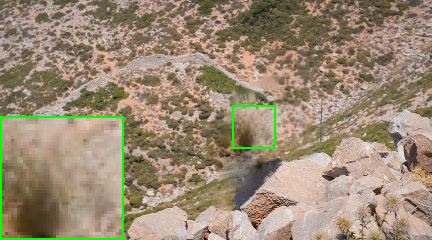} &
		\includegraphics[width=34.15mm]{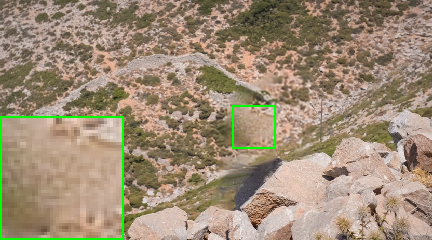} &
		\includegraphics[width=34.15mm]{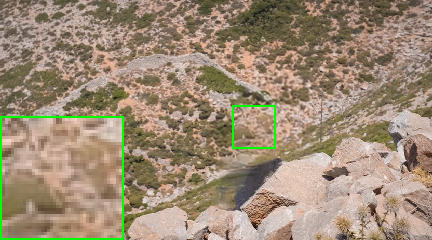} \\

		\includegraphics[width=34.15mm]{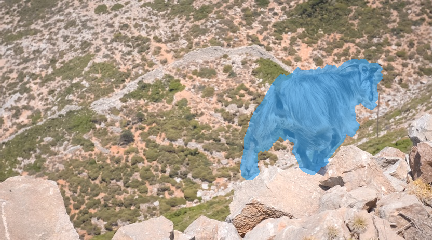} &
		\includegraphics[width=34.15mm]{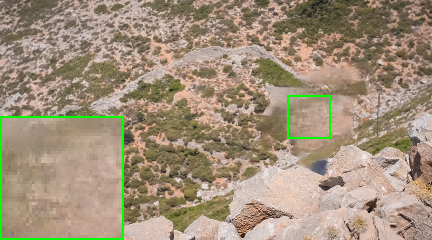} &
		\includegraphics[width=34.15mm]{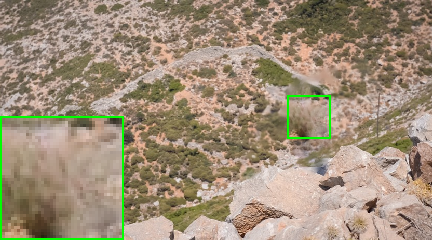} &
		\includegraphics[width=34.15mm]{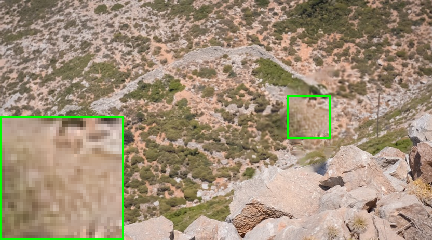} &
		\includegraphics[width=34.15mm]{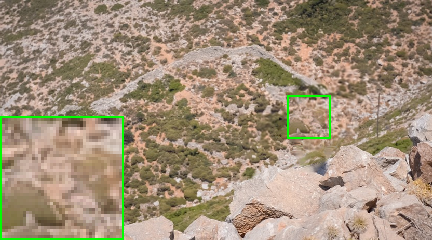} \\

		\includegraphics[width=34.15mm]{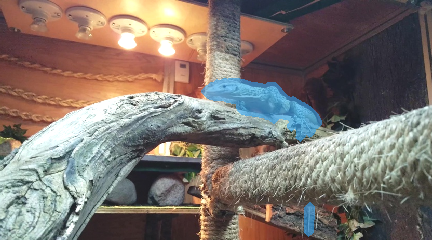} &
		\includegraphics[width=34.15mm]{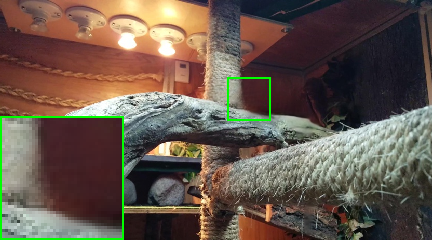} &
		\includegraphics[width=34.15mm]{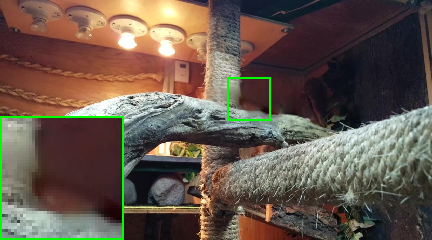} &
		\includegraphics[width=34.15mm]{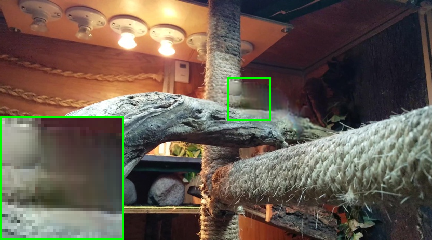} &
		\includegraphics[width=34.15mm]{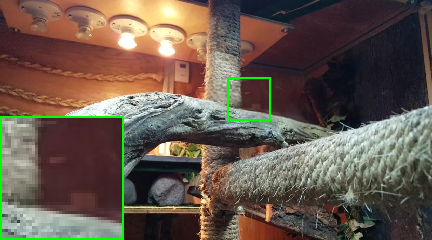} \\

		\includegraphics[width=34.15mm]{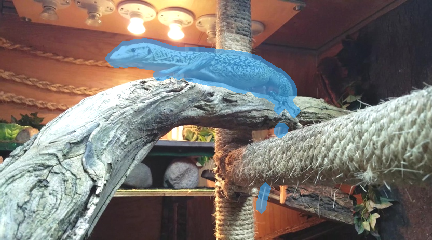} &
		\includegraphics[width=34.15mm]{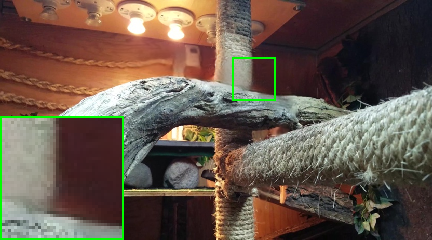} &
		\includegraphics[width=34.15mm]{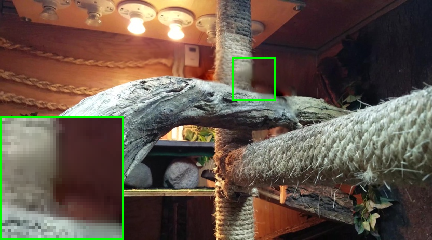} &
		\includegraphics[width=34.15mm]{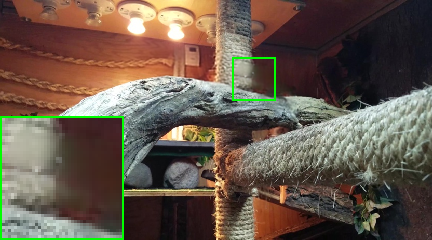} &
		\includegraphics[width=34.15mm]{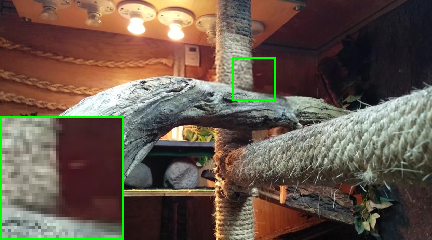} \\

            \includegraphics[width=34.15mm]{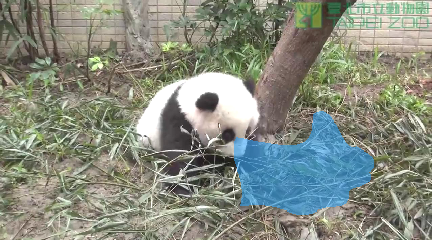} &
		\includegraphics[width=34.15mm]{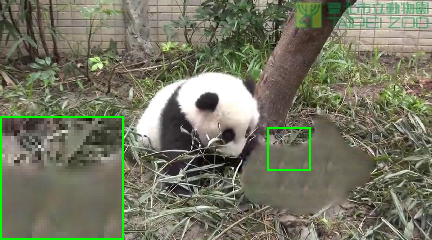} &
		\includegraphics[width=34.15mm]{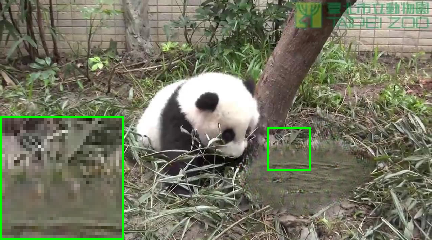} &
		\includegraphics[width=34.15mm]{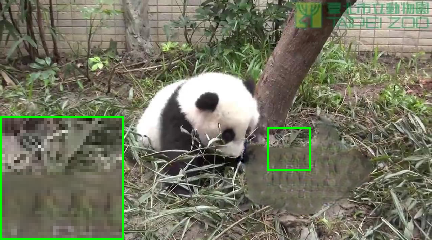} &
		\includegraphics[width=34.15mm]{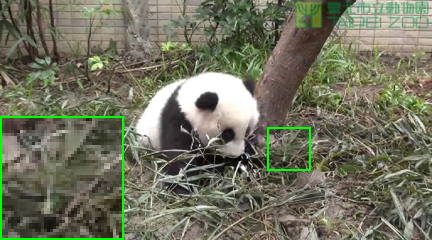} \\

            \includegraphics[width=34.15mm]{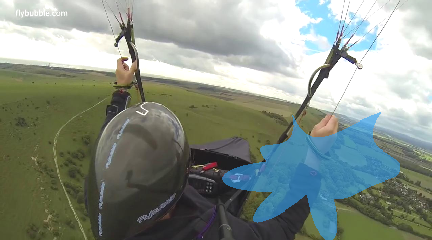} &
		\includegraphics[width=34.15mm]{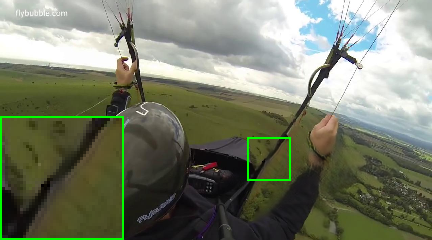} &
		\includegraphics[width=34.15mm]{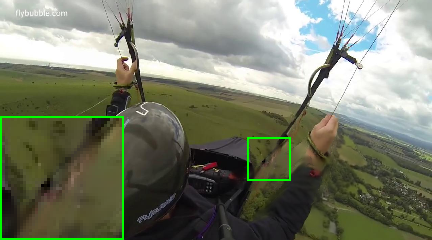} &
		\includegraphics[width=34.15mm]{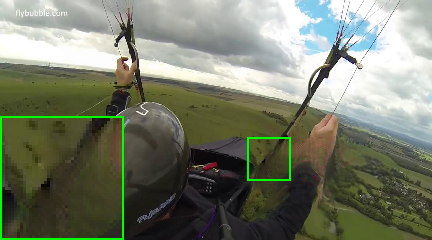} &
		\includegraphics[width=34.15mm]{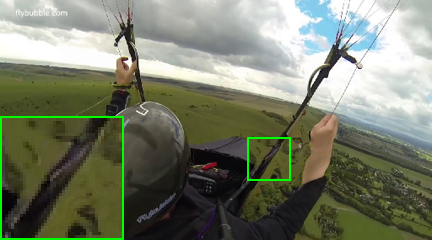} \\

		(a) Input & (b) ProPainter~\cite{zhou2023propainter} & (c) FGT~\cite{zhang2022flow} & (d) ECFVI~\cite{kang2022error}  & (e) Ours\\

	\end{tabular}
	\caption{Qualitative comparisons with SOTA video inpainting methods. Best viewed in PDF with zoom. Please refer to the supplementary video for a comprehensive comparison. }
	\label{visual_compare}

\vspace{-1.5em}
\end{figure*}

\section{Experiments}

\subsection{Experimental Settings}
\subsubsection{Dataset} Following common practice, we adopt two widely used video object segmentation datasets, namely YouTube-VOS~\cite{xu2018youtube} and DAVIS~\cite{perazzi2016benchmark}, for evaluation. YouTube-VOS comprises 3471, 474, and 508 video clips for training, validation, and testing respectively. DAVIS consists of $150$ video clips, with 60 for training, 30 for validation and 90 for testing. Since we do not need any training data, we follow the same setting as FuseFormer~\cite{liu2021fuseformer} and E2FGVI~\cite{li2022towards} to use test sets from YouTube-VOS and DAVIS for quantitative and qualitative evaluations. For the testing masks, we adopt the same stationery masks as E2FGVI~\cite{li2022towards} for computing objective metrics and object shape masks for qualitative comparisons. Our code will be made publicly available at: \href{https://vious.github.io/projects/VipDiff}{https://vious.github.io/projects/VipDiff}.

\vspace{-1.0em}
\subsubsection{Quantitative Metrics} For a fair comparison, we select PSNR, SSIM~\cite{wang2004image}, VFID~\cite{NEURIPS2018_VFID}, and flow warping error $\text{E}_{warp}$~\cite{lai2018learning_ewarp} to evaluate the performance of our method quantitatively with other state-of-the-art video inpainting methods. 

\vspace{-1.0em}
\subsubsection{Implementation Details} We adopt the pre-trained image-level diffusion model from LDMs~\cite{rombach2022high}, and the number of reverse optimization steps is set to 50, with an adaptive learning rate initialized at $\eta_0=0.01$, and $\eta_t=0.9\eta_{t-1}$ for optimizing the noise. We set the weight for the regularization term to $\gamma=0.001$. We adopt the pre-trained RAFT provided by ~\cite{kang2022error} as our flow completion model, and also utilize their error compensation model for reducing color discrepancy in pixel propagation. The whole generating process is performed on a sinlge GTX 3090 GPU. For a corrupted video with $100$ frames, our method takes about 4 to 6 minutes to complete a video, which is also efficient since our \modelname does not require any training time and training data. 

To select the first frame $x^{k}_{0}$ for constrained generation, we warp the masks of each corrupted frames using the completed flows, and then select the frame which holds the most overlapping mask region as the starting frame.

\subsection{Main Comparisons}
\subsubsection{Quantitative Results} 
We compute the quantitative results on the YouTube-VOS and DAVIS datasets, and compare with existing state-of-the-art methods, including LGTSM~\cite{chang2019learnable}, VINet~\cite{kim2019deep}, DFVI~\cite{xu2019deep}, CAP~\cite{lee2019copy}, STTN~\cite{zeng2020learning}, FGVC~\cite{gao2020flow}, FuseFormer~\cite{liu2021fuseformer}, E2FGVI~\cite{li2022towards}, ProPainter~\cite{zhou2023propainter}, FGT~\cite{zhang2022flow} and ECFVI~\cite{kang2022error}. Following E2FGVI and ECFVI, we test all the frames at the resolution of $432\times240$. As shown in Tab.~\ref{metric_compare}, although our \modelname does not require any training on those datasets, it achieved the most highest metric scores compared with other SOTA methods. As for the video related metrics, we also achieved the highest $\text{E}_{warp}$ and VFID both on YouTube-VOS and on DAVIS datasets. This demonstrates the superior performance of \modelname.

\noindent\textbf{Inference time.} We compared the running time per image on the DAVIS validation dataset using a single RTX 3090 GPU and Intel 4214R CPU, as shown in Table~\ref{metric_compare} (last column). On average, our \modelname is only about $1.4$s slower than optical-flow guided methods like ECFVI, but providing better visual quality. Diffusion models themselves are time-consuming (LDM takes about 4.3s just for inference), our framework can do object removal for a video sequence about 100 frames in about 4 minutes, which is time-efficient for video inpainting. For inpainting task, users focus more on visual quality, though feed-forward models like E2FGVI or Propainter takes less time, they produce severe artifacts. With the faster modern GPU techniques and better acceleration algorithms of diffusion models in future, our \modelname is possible to close the speed gap to those feed-forward models, while providing sharper and high-detailed results.

\begin{figure}[bt]
	\small
	\setlength{\tabcolsep}{1.0pt}
	\centering
	\begin{tabular}{cccc}

        \rotatebox{90}{(a) Input} & 
        \includegraphics[width=27mm]{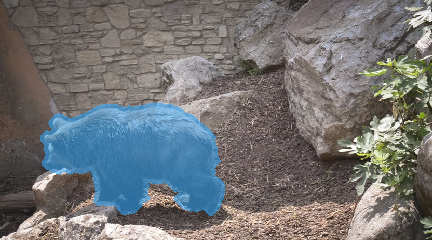} &
        \includegraphics[width=27mm]{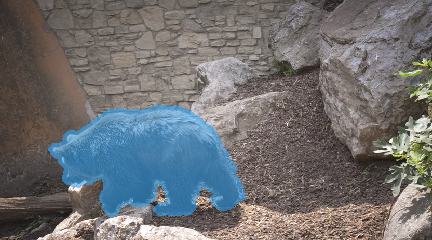} &
        \includegraphics[width=27mm]{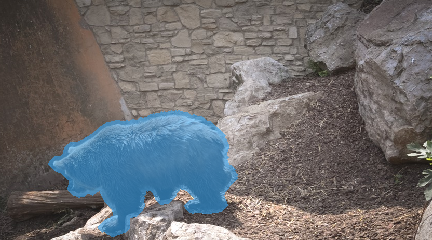}  \\

        \rotatebox{90}{(b)  LDM} &
        \includegraphics[width=27mm]{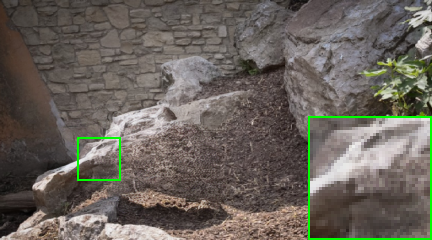} & 
        \includegraphics[width=27mm]{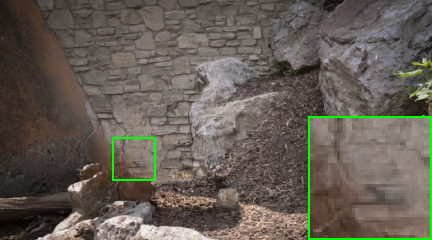} & 
        \includegraphics[width=27mm]{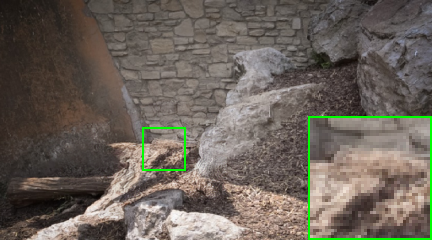} \\

        \rotatebox{90}{(c) LDM+PP} &
        \includegraphics[width=27mm]{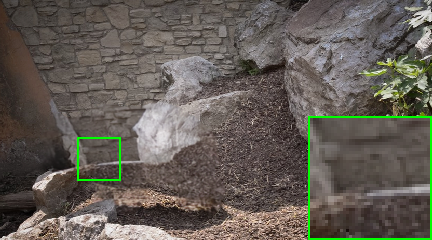} &
        \includegraphics[width=27mm]{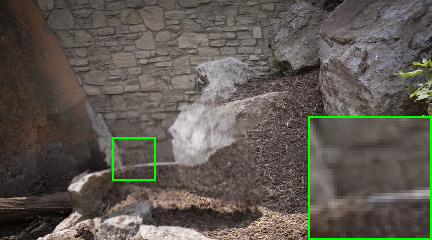} &
        \includegraphics[width=27mm]{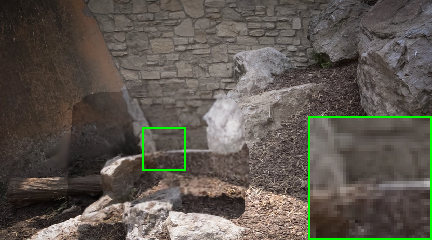}\\

        \rotatebox{90}{(d) w/o Opt} &
        \includegraphics[width=27mm]{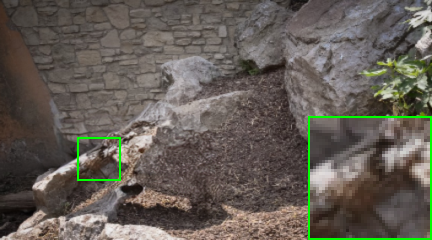} &
        \includegraphics[width=27mm]{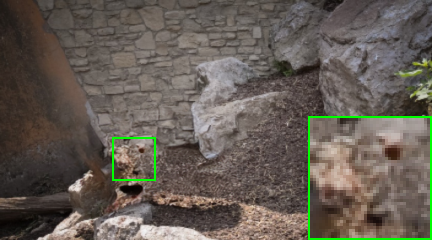} &
        \includegraphics[width=27mm]{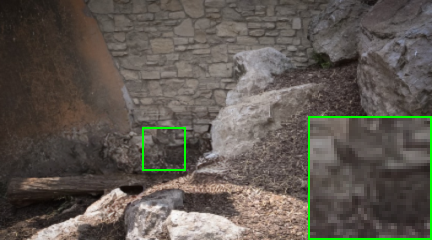}\\

        \rotatebox{90}{(e) Ours} &
        \includegraphics[width=27mm]{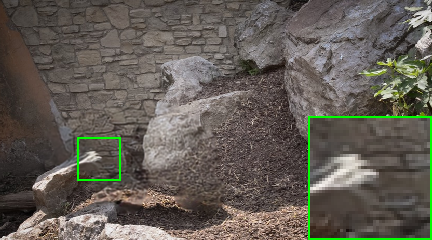} &
        \includegraphics[width=27mm]{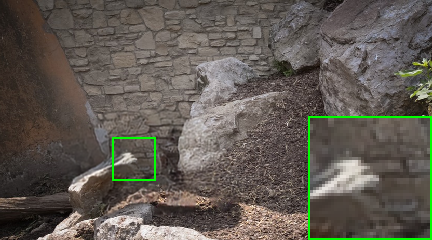} &
        \includegraphics[width=27mm]{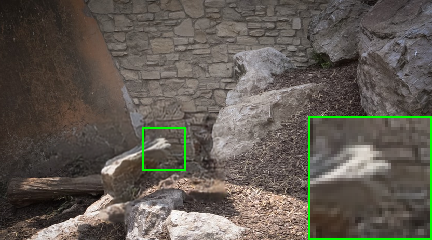} \\

	\end{tabular}
	\caption{Ablation study on different variants. From top row to the bottom, they are (a) input, (b) output of LDM, (c) LDM combined pixel propagation process, (d) our framework without reverse noise optimization, and (e) ours.}
	\label{ablation}

\vspace{-2.0em}
\end{figure}

\subsubsection{Qualitative Results}
For qualitative comparisons, we select three recent SOTA methods, namely ProPainter~\cite{zhou2023propainter}, FGT~\cite{zhang2022flow}, and ECFVI~\cite{kang2022error}, to conduct visual comparisons. We select two videos with object-shaped masks (top 4 rows in Fig.~\ref{visual_compare}) and two videos with stationery masks (bottom 2 rows in Fig.~\ref{visual_compare}). One can see that for object-removal task (top 4 rows), our \modelname can generate both sharp and temporal-coherent results, while the competing methods fail to generate temporal-coherent and sharp contents in the mask center. As for the stationery masks (bottom 2 rows), our \modelname largely surpasses the SOTA video inpainting methods since they cannot find useful contents in the reference frames, while our training-free framework can generate meaningful details in the masked region. We refer readers to our video results in the supplementary material for a better evaluation of the temporal-coherent performance of our \modelname.

\subsection{Ablation Study}
We conduct an ablation study on the DAVIS dataset with the following variants: ($\romannumeral1$) direct output of LDM~\cite{rombach2022high} inpainting model for all corrupted frames, ($\romannumeral 2$) iteratively select a corrupted frame to inpaint using LDM and use completed flows to propagate pixels to other neighbouring frames, until there is no invalid mask, ($\romannumeral 3$) without noise optimization in the reverse diffusion process (i.e., use optical flow to guide pixel propagation first and then randomly sample a fixed Gaussian noise for inpainting all the corrupted frames). We denote  these three variants as 
`LDM', 
`LDM+PP', and 
`w/o Opt' respectively.

We show the qualitative results of the ablation study in Fig.~\ref{ablation}, with three consecutive frames randomly selected from the `bear' video. One can clearly see that when directly applying LDM inpainting model without considering any temporal prior, though the generated results are sharp for every frame, they are not temporal-coherent (see Fig.~\ref{ablation}(b)). For LDM+PP, the inside inpainted contents may be temporal-coherent, but directly warping large patches yield severe color discrepancy issues due to the brightness of different frames. From Fig.~\ref{ablation}(d), one can see that removing the noise optimization step and using fixed Gaussian noise for all corrupted frames would also generate different contents inside the mask center. Our \modelname, on the other hand, can generate both sharp and temporal-coherent contents, demonstrating its effectiveness. 

We show the quantitative results of the ablation study in Tab.~\ref{ablation_metrics}. One can note that our full framework yielded the highest quantitative performance over other variants. LDM achieved the worst results in all metrics since it does not consider any temporal coherence. LDM+PP improved over LDM by adopting optical flow for pixel propagation, but it faces color discrepancy and structure incoherent issues. Removing the noise optimization step (w/o Opt) largely decreased the video-related quantitative performance. Furthermore, we conducted additional experiments by performing noise optimized reverse diffusion process for every single frame. However, we observed that this approach was considerably slow and resulted in significant frame flickering issues. Hence, we omit this result in comparison. 


         
         
    

\begin{table}[bt]
	\centering
        \setlength{\abovecaptionskip}{-0.1em}
	\caption{Ablation studies on different variants.}
	\label{ablation_metrics}
	\begin{tabular}{c|c|c|c|c}
		\hline  
		
		\hline

        Variant & PSNR$\uparrow$ & SSIM$\uparrow$ & VFID$\downarrow$ & $\text{E}_{warp}$ $\downarrow$ \\
  
	\hline

        LDM &  32.64 & 0.9673 & 0.206 & 0.1833 \\
        
        \hline

        LDM+PP &  33.14 & 0.9701 & 0.133 & 0.1379 \\

        \hline

        w/o Opt &  32.80 & 0.9698 & 0.187 & 0.1620 \\

        \hline

        Ours & {34.23} & {0.9745} & {0.102} & {0.1280} \\ 
		
		\hline
		
		\hline
		
	\end{tabular} 

\vspace{-0.5em}
\end{table}

\begin{figure}[bt]
	\small
	\setlength{\tabcolsep}{0.5pt}
	\centering
	\begin{tabular}{ccccc}

        \includegraphics[width=16.4mm]{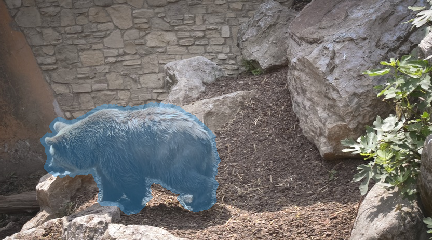} &
        \includegraphics[width=16.4mm]{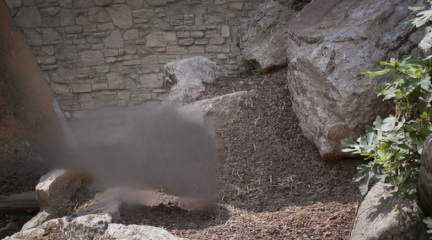} &
        \includegraphics[width=16.4mm]{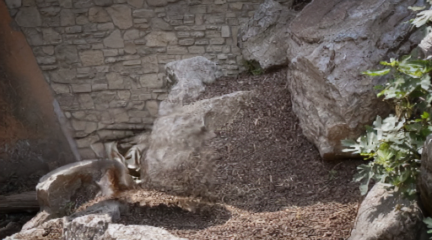} &
        \includegraphics[width=16.4mm]{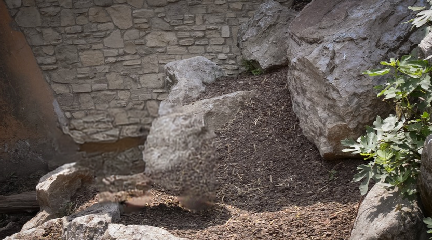} & 
        \includegraphics[width=16.4mm]{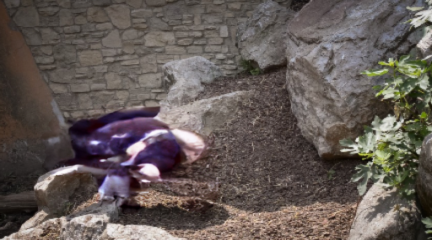} \\ 

        Input & $\gamma=0.1$ & $\gamma=0.01$ &  $\gamma=0.001$ & $\gamma=0$ \\

	\end{tabular}
        \vspace{-1.0em}
	\caption{Ablation study on regularization weight $\gamma$. }
	\label{ablation_gamma}

\vspace{-1.5em}
\end{figure}

\noindent \textbf{Ablation on $\gamma R(z)$.} We conducted experiments on different weights of $\gamma$, ranging from $0.1$ to $0$, as illustrated in Fig~\ref{ablation_gamma}. As the weight increases, we observed the blurriness and spatial distortions. This can be attributed to the heavier burden imposed by large Gaussian constraints on the later reverse diffusion steps, resulting in a higher magnitude of noise level. On the other hand, setting $\gamma=0$, implying the absence of Gaussian constraints, may lead to the occurrence of artifacts. This is likely because the reconstruction loss induces rapid changes in the noise distribution during the early denoising steps, where it should still resemble Gaussian noise. Based on our experiments, we recommend selecting a slightly small $\gamma$ ranging from $0.01$ to $0.001$ (avoiding values that are too large or too small, such as 0) to achieve satisfactory results. 

We refer readers to our supplementary material for visual comparisons in videos and more ablation study, which more clearly delineate the strength of our method. Further, our training-free framework possesses strong generalization capability. It can be applied to any pre-trained image-level unconditional denoising diffusion models for video inpainting task. We provide video results by adopting other diffusion models in the suppl.

\section{Conclusion}
In this paper, we have proposed the first training-free framework, named \modelname, that effectively tames a pre-trained image-level diffusion model for the video inpainting task. 
By introducing an optical flow guided reverse noise optimization process, our framework successfully generates sharp and temporally coherent video completion results.
Our \modelname further allows for providing diverse video outputs, and it also saves significant amount of efforts in collecting extensive video data for training a video inpainting diffusion model. Experiments have shown that our method achieves the state-of-the-art performance on benchmark datasets, and generates superior video completion results than other competing methods.

\noindent \textbf{Acknowledgement} \
This work is partially supported by the Hong Kong Research Grants Council - General Research Fund (Grant No.:~17211024) and HKU Seed Fund for Basic Research.

\clearpage
{\small
\bibliographystyle{ieee_fullname}
\bibliography{egbib}
}

\end{document}